%% file: main.tex
\documentclass[conference]{IEEEtran}
\usepackage[utf8]{inputenc}
\IEEEoverridecommandlockouts

\usepackage{cite}
\usepackage{amsmath,amssymb,amsfonts}
\usepackage{algorithmic}
\usepackage{graphicx}
\usepackage{textcomp}
\usepackage{xcolor}
\usepackage{soul}
\usepackage{booktabs}
\usepackage{threeparttable}
\usepackage{hyphenat}
\usepackage{threeparttable}
\usepackage{stfloats}
\usepackage{multirow}
\usepackage{colortbl}
\usepackage{makecell}
\usepackage{amssymb}
\usepackage{svg}
\usepackage{pifont}
\usepackage{orcidlink}
\usepackage{caption}
\usepackage{cleveref}
\usepackage{circledsteps}
\usepackage[acronym]{glossaries}
\input{text/acronyms.tex}
\newcommand{\cmark}{\checkmark}
\newcommand{\xmark}{\ding{55}} 

\newcommand*\haeggli{\Circled[inner color=white, outer color=white, fill color=green!60!black]{\cmark}}
\newcommand*\chruezli{\Circled[inner color=white, outer color=white, fill color=red,]{\,\xmark\,}}
\newcommand*\welleli{\Circled[inner color=white, outer color=white, fill color=orange, inner ysep=6pt]{$\approx$}}

\def\BibTeX{{\rm B\kern-.05em{\sc i\kern-.025em b}\kern-.08em
    T\kern-.1667em\lower.7ex\hbox{E}\kern-.125emX}}

\renewenvironment{thebibliography}[1]{%
  \begin{oldthebibliography}{#1}%
    \setlength{\itemsep}{0pt}%
    \setlength{\parskip}{0pt}%
}{\end{oldthebibliography}}

\glsenableentrycount

\begin{document}
\pagestyle{plain}
\title{BioTrain: Sub-MB, Sub-50mW On-Device Fine-Tuning for Edge-AI on Biosignals\thanks{\textcopyright\,2026 IEEE. Accepted at the 2026 International Joint Conference on Neural Networks (IJCNN), part of the IEEE World Congress on Computational Intelligence (WCCI). Personal use of this material is permitted. Permission from IEEE must be obtained for all other uses, in any current or future media, including reprinting/republishing this material for advertising or promotional purposes, creating new collective works, for resale or redistribution to servers or lists, or reuse of any copyrighted component of this work in other works.}}

\input{text/author} 

\maketitle
\glsresetall

\input{text/00-abstract}


\input{text/01-introduction}

\input{text/02-background}

\input{text/03-methods}

\input{text/04-results}

\input{text/05-conclusions}

\section*{Acknowledgment}
This work has received funding from the Swiss State Secretariat for Education, Research, and Innovation (SERI) under the SwissChips initiative.

\bibliographystyle{IEEEtran}
\bibliography{references_final}
\end{document}

%% file: text/acronyms.tex
\newacronym{AI}{AI}{Artificial Intelligence}
\newacronym{ml}{ML}{machine learning}
\newacronym{bn}{BN}{Batch Normalization}
\newacronym{gn}{GN}{Group Normalization}
\newacronym{ln}{LN}{Layer Normalization}
\newacronym{in}{IN}{Instance Normalization}
\newacronym{DL}{DL}{Deep Learning}
\newacronym{DNN}{DNN}{Deep Neural Network}
\newacronym[longplural={Application-Specific Integrated Circuits}]{ASIC}{ASIC}{Application-Specific Integrated Circuit}
\newacronym[longplural={Graphics Processing Units}]{GPU}{GPU}{Graphics Processing Unit}
\newacronym{TPU}{TPU}{Tensor Product Unit}
\newacronym{CPU}{CPU}{Central Processing Unit}
\newacronym[longplural={Instruction Set Architectures}]{ISA}{ISA}{Instruction Set Architecture}
\newacronym[plural=TCNs, firstplural={Temporal Convolutional Neural Networks (TCNs)}]{TCN}{TCN}{Temporal Convolutional Neural Network}
\newacronym[plural=CNNs, firstplural={Convolutional Neural Networks (CNNs)}]{CNN}{CNN}{Convolutional Neural Network}
\newacronym{MHSA}{MHSA}{Multi-Head Self-Attention}
\newacronym{MHA}{MHA}{Multi-Head Attention}
\newacronym[longplural={Reduced Instruction Set Computers}]{RISC}{RISC}{Reduced Instruction Set Computer}
\newacronym{ARM}{ARM}{Advanced RISC Machine}
\newacronym{SSR}{SSR}{Stream Semantic Register}
\newacronym{HW}{HW}{Hardware}
\newacronym{WL}{WL}{Workload}
\newacronym{ONNX}{ONNX}{Open Neural Network Exchange}
\newacronym{NAS}{NAS}{Neural Architecture Search}
\newacronym{PULP}{PULP}{Parallel Ultra Low Power}
\newacronym{AXI}{AXI}{Advanced eXtensible Interface}
\newacronym[longplural={Micro-Controller Units}]{MCU}{MCU}{Micro-Controller Unit}
\newacronym{ODL}{ODL}{On-Device Learning}
\newacronym{SGD}{SGD}{Stochastic Gradient Descent}
\newacronym{FLOPS}{FLOPS}{Floating Point Operations Per Second}
\newacronym{SNN}{SNN}{Spiking Neural Network}
\newacronym{KWS}{KWS}{Keyword Spotting}
\newacronym{IIS}{IIS}{Integrated Systems Laboratory}
\newacronym[longplural={Large Language Models}]{LLM}{LLM}{Large Language Model}
\newacronym[longplural={Systems-on-Chip}]{SoC}{SoC}{System-on-Chip}
\newacronym{NLP}{NLP}{Natural Language Processing}
\newacronym{SotA}{SotA}{State of the Art}
\newacronym{CV}{CV}{Computer Vision}
\newacronym{MAC}{MAC}{Multiply-Accumulate}
\newacronym{IoT}{IoT}{Internet of Things}
\newacronym{SIMD}{SIMD}{Single Instruction Multiple Data}
\newacronym{PTQ}{PTQ}{Post-Training Quantization}
\newacronym{QAT}{QAT}{Quantization Aware Training}
\newacronym{EEG}{EEG}{Electroencephalogram}
\newacronym{RAW}{RAW}{Read-After-Write}
\newacronym{WRL}{WRL}{Weight-Reuse Linear}
\newacronym{IRL}{IRL}{Input-Reuse Linear}
\newacronym{LWT}{LWT}{Layer-Wise Tiling}
\newacronym{DFT}{DFT}{Depth-First Tiling}
\newacronym{MQA}{MQA}{Multi-Query Attention}
\newacronym{GQA}{GQA}{Grouped-Query Attention}
\newacronym[longplural={Vision Transformers}]{ViT}{ViT}{Vision Transformer}
\newacronym{GELU}{GELU}{Gaussian Error Linear Unit}
\newacronym{GEMM}{GEMM}{GEneral Matrix Multiplication}
\newacronym{HPC}{HPC}{High-Performance Computing}
\newacronym[longplural={Direct Memory Accesses}]{DMA}{DMA}{Direct Memory Access}
\newacronym{PE}{PE}{Processing Element}
\newacronym{FPU}{FPU}{Floating-Point Unit}
\newacronym{RAM}{RAM}{Random-Access Memory}
\newacronym{SRAM}{SRAM}{Static Random-Access Memory}
\newacronym{NE16}{NE16}{Neural Engine 16-channels}
\newacronym{FWSA}{FWSA}{Fused-Weight Self-Attention}
\newacronym{TQT}{TQT}{Trained Quantization Threshold}
\newacronym{ReLU}{ReLU}{Rectified Linear Unit}
\newacronym{ML}{ML}{Machine Learning}
\newacronym{SLM}{SLM}{Small Language Model}
\newacronym{FC}{FC}{Fully-Connected}
\newacronym{MLP}{MLP}{Multi-Layer Perceptron}

\newacronym{lora}{LoRA}{Low-Rank Adaptation}
\newacronym{riscv}{RISC-V}{Reduced Instruction Set Computer V}
\newacronym{onnx}{ONNX}{Open Neural Network Exchange}
\newacronym{soc}{SoC}{System-on-Chip}
\newacronym{gemm}{GEMM}{General Matrix Multiplication}
\newacronym{dma}{DMA}{Direct Memory Access}
\newacronym{l1}{L1}{Level-1 Tightly-Coupled Data Memory (TCDM)}
\newacronym{l2}{L2}{On-Chip SRAM}
\newacronym{l3}{L3}{External Memory (e.g., HyperRAM)}
\newacronym{TCDM}{TCDM}{Tightly-Coupled Data Memory}
\newacronym{mcu}{MCU}{Microcontroller Unit}
\newacronym{sram}{SRAM}{Static Random-Access Memory}
\newacronym{bp}{BP}{Backpropagation}
\newacronym{ir}{IR}{Intermediate Representation}

\newacronym{cl}{CL}{Continual Learning}
\newacronym{er}{ER}{Experience Replay}
\newacronym{adamw}{AdamW}{Adaptive Moment Estimation with Weight Decay}
\newacronym{sgd}{SGD}{Stochastic Gradient Descent}

\newacronym{eeg}{EEG}{Electroencephalogram}
\newacronym{eog}{EOG}{Electrooculogram}
\newacronym{emg}{EMG}{Electromyogram}
\newacronym{bci}{BCI}{Brain-Computer Interface}

\newacronym{loso}{LOSO}{Leave-One-Subject-Out}

\newacronym{lp}{LP}{Linear Probing}
\newacronym{fullft}{Full-FT}{Full Backpropagation Fine-Tuning}
\newacronym{edgeft}{Edge-FT}{Edge-Optimized Fine-Tuning}
\newacronym{noft}{No-FT}{Zero-Shot / No Fine-Tuning}

%% file: text/author.tex

\author{
\IEEEauthorblockN{
    Run Wang\IEEEauthorrefmark{1}~\orcidlink{0000-0002-1011-6432},
    Victor J.B. Jung\IEEEauthorrefmark{1}~\orcidlink{0009-0001-7462-3468},
    Philip Wiese\IEEEauthorrefmark{1}~\orcidlink{0009-0001-7214-2150}, 
    Sebastian Frey\IEEEauthorrefmark{1}~\orcidlink{0009-0000-6948-4363},
    Giusy Spacone\IEEEauthorrefmark{1}, \\
    Francesco Conti\IEEEauthorrefmark{2}~\orcidlink{0000-0002-7924-933X}, 
    Alessio Burrello\IEEEauthorrefmark{3}~\orcidlink{0000-0002-6215-8220},
    Luca Benini\IEEEauthorrefmark{1}\IEEEauthorrefmark{2}~\orcidlink{0000-0001-8068-3806}
}
\IEEEauthorblockA{
    \IEEEauthorrefmark{1}\textit{Integrated Systems Laboratory (IIS), ETH Zurich}, Switzerland \\
    \IEEEauthorrefmark{2}\textit{Department of Electrical, Electronic and Information Engineering (DEI), University of Bologna}, Italy \\
    \IEEEauthorrefmark{3}\textit{Department of Control and Computer Engineering (DAUIN), Politecnico di Torino}, Italy
}
\IEEEauthorblockA{
    \{runwang, jungvi, wiesep, sebastian.frey, gspacone, lbenini\}@iis.ee.ethz.ch, f.conti@unibo.it, alessio.burrello@polito.it
  }
    \vspace{-0.5cm}
}

%% file: text/00-abstract.tex
\begin{abstract}
Biosignals exhibit substantial cross-subject and cross-session variability, inducing severe domain shifts that degrade post-deployment performance for small, edge-oriented AI models. On-device adaptation is therefore essential to both preserve user privacy and ensure system reliability. 
However, existing sub-100\,mW \gls{mcu}-based wearables platforms \textcolor{black}{can only support} shallow or sparse adaptation schemes due to the prohibitive memory footprint and computational cost of full \gls{bp}. 
In this paper, we propose BioTrain, a framework enabling full-network fine-tuning of state-of-the-art biosignal models under milliwatt-scale power and sub-megabyte memory constraints. 
We validate BioTrain using both offline and on-device benchmarks on \gls{eeg} and \gls{eog} datasets, covering Day-1 new-subject calibration and longitudinal adaptation to signal drift. Experimental results show that full-network fine-tuning achieves accuracy improvements of up to 35\% over non-adapted baselines and outperforms last-layer update by approximately 7\% during new-subject calibration. 
Furthermore, on the GAP9 \gls{mcu} platform, BioTrain enables efficient on-device training throughput of 17 samples/s for \gls{eeg} and 85 samples/s for \gls{eog} models with a power envelope below 50\textcolor{black}{\,mW}.
In addition, BioTrain's efficient memory allocator and network topology optimization enable the use of a large batch size, thereby reducing peak memory usage. For a completely on-chip \gls{bp} on GAP9, BioTrain reduces the memory footprint by 8.1$\times$ from 5.4\,MB to 0.67\,MB\textcolor{black}{compared to conventional full-network fine-tuning using batch normalization with batch size 8}.
\end{abstract}

%% file: text/01-introduction.tex
\vspace{-0.5em}
\section{Introduction}
\glsresetall

Biosignal-driven edge-AI is rapidly emerging in transformative wearable applications, ranging from \gls{bci} to longitudinal health monitoring~\cite{li_survey_2025}. \textcolor{black}{B}iosignals exhibit substantial inter-subject variability and non-stationarity during long-term use, driven by factors such as fluctuating electrode--skin impedance, perspiration, and sensor displacement~\cite{yang_multi-day_2025, li_survey_2025}. These effects induce distribution shifts between training and deployment, degrading performance in wearable-grade, compact AI models, under both cross-subject transfer and long-term temporal drift.
To address this challenge, on-device adaptation enables continuous model updates based on biosignals acquired in situ, allowing personalization under non-stationary sensing conditions~\cite{zhu_-device_2024}. 
When cloud-trained models are deployed on edge devices for biosignal-based wearables, on-device adaptation typically consists of an initial calibration phase for new users, followed by continuous adaptation to long-term signal drift.
By avoiding reliance on remote servers, this step naturally preserves data privacy and supports stable, low-latency operations that do not depend on a remote connection. 

However, the practical deployment of on-device training techniques is constrained by limited computational, memory, and energy resources in wearable devices. These constraints make robust deep adaptation difficult and have traditionally forced aggressive memory-reduction strategies, such as sparse updates or last-layer training, also known as \gls{lp}.
Considering both on-device adaptation phases, restricting training to only partially update the network's weights has been shown to be insufficient to handle substantial signal variations~\cite{mei_ultra-low_2025}. Enabling deeper on-device \gls{bp} is therefore desirable, but it is fundamentally constrained by the large memory footprint of \gls{bp} and the complexity of executing its computation under tight resource budgets. In the absence of ecosystem and compiler support for automated memory orchestration, such deployments rely heavily on manual optimization, severely limiting the scalability and robustness of on-device adaptation.

To address these challenges, we propose BioTrain, a compiler framework that automatically generates efficient bare-metal C code to perform on-device training on resource-constrained \gls{mcu} platforms, enabling full-network \gls{bp} under tight memory budgets. BioTrain is built on Deeploy, a domain-specific compiler originally developed for energy-efficient inference on \glspl{mcu}~\cite{scherer_deeploy_2024}, and extends its inference-oriented compilation pipeline to support training workloads. In particular, BioTrain leverages Deeploy's static memory allocation and tiling mechanisms and extends them to gradient computations by integrating optimized gradient kernels from PULPTrainLib~\cite{orailoglu_pulp-trainlib_2022}, enabling efficient \gls{bp} for long time-series workloads. 
The main contributions of this work are as follows:

\begin{itemize}
\item We develop BioTrain, a deployment framework that \textcolor{black}{support} \gls{bp} code generation on resource-constrained devices by integrating and optimizing \gls{CNN} gradient kernels tuned for tiling from PULPTrainLib. 

\item We demonstrate two real-world on-device biosignal adaptation scenarios on \gls{eeg}~\cite{mei_ultra-low_2025} and \gls{eog}~\cite{frey_gapses_2025} datasets. By modifying the topology of the baseline \gls{CNN} architectures to support on-chip mini-batch processing, BioTrain achieves up to an 8$\times$ peak activation memory reduction, from 5.4\,MB to 0.67\,MB, \textcolor{black}{with respect to conventional full-network fine-tuning using \gls{bn} with batch size 8}, enabling full-network on-chip \gls{bp} while maintaining comparable or improved performance under both Day-1 new-user calibration and longitudinal adaptation.

\item We evaluate on GAP9 \gls{mcu} showing up to 35\% accuracy improvement over non-adapted baselines and 7\% over \gls{lp} across both Day-1 Calibration and Longitudinal Adaptation scenarios. BioTrain achieves full \gls{bp} training throughput of 17 samples/s (\gls{eeg}) and 85 samples/s (\gls{eog}) within a 50\,mW power envelope. Energy analysis shows that a standard 320\,mAh battery supports 211 (\gls{eeg}) or 951 (\gls{eog}) complete training sessions with a training configuration of 40 epochs over 200 samples, enabling practical repeated on-device personalization and adaptation.

\end{itemize}

\textcolor{black}{BioTrain is released open source at \texttt{https://github.com/pulp-platform/Deeploy}.}

%% file: text/02-background.tex
\section{Background and Related Works}
\subsection{On-device Adaptation}
On-device adaptation has become a key capability for edge biosignal-based AI systems operating in non-stationary conditions, where user behavior, sensor placement, and environmental factors induce distribution shift after deployment.
\textcolor{black}{
Prior work spans (i) supervised personalization via parameter-efficient fine-tuning (e.g., low-rank adaptation or only updating selected layers)~\cite{han2024parameter}, and (ii) inference-time unsupervised/semi-supervised test-time adaptation (TTA) based on self-supervision (e.g., entropy minimization, self-distillation, pseudo-labeling), typically updating only a small subset of parameters to fit edge constraints~\cite{song2023ecotta,jia2024tinytta}.
Although \gls{bp}-free or statistics-only updates reduce memory usage, they inherently limit adaptation capacity when task-specific supervision is available, as is common during short calibration sessions on biosignal-based wearable devices.
BioTrain therefore targets true gradient-based on-device learning to match offline optimization and generalizing across model architectures without hand-crafted update rules.
}

\subsection{AI Deployment Compiler Frameworks}
Edge deployment differs from general-purpose DNN stacks (e.g., PyTorch/JAX) due to tight on-chip memory and limited runtime support.
\textcolor{black}{
Commercial and open-source compiler toolchains provide efficient \cgls{mcu}-centric inference (e.g., TFLM, X-CUBE-AI, NXP's eIQ, and compiler ecosystems such as MLIR/TVM for lowering and code generation)~\cite{Lin2023TinyML}. On the academic side, Deeploy~\cite{scherer_deeploy_2024} and MATCH~\cite{amine_hamdi_match_2025} demonstrate compiler-driven tiling and static memory allocation for memory-constrained inference using optimized operator libraries such as PULP-NN. TrainDeeploy~\cite{wang2026traindeeploy} extends Deeploy to adapter-style PEFT of small transformers at the edge, but no existing stack supports end-to-end full-network on-device training. Hence, BioTrain extends Deeploy from inference to on-device adaptation, enabling end-to-end \gls{bp} for \glspl{CNN}.
}

\subsection{On-device Training Frameworks}

\begin{table}[t]
\centering
\caption{Representative on-device training frameworks.}
\label{tab:related_ondevice_training}
\scriptsize
\resizebox{\columnwidth}{!}{%
\begin{tabular}{lllclc}
\toprule
Work & Algorithm & Depth & BS $>1$ 
& Acc. (rel.)$^{\dagger}$ & Compiler \\
\midrule
TinyOL~\cite{ren2021tinyol} 
& Last-layer 
& LP 
& \chruezli
& $\downarrow$ 5--10\% 
& \chruezli~ \\

MiniLearn~\cite{profentzas2022minilearn} 
& Prune + FT 
& Sparse 
& \haeggli
& $\downarrow$ 2--10\% 
& \chruezli~ \\

TTE~\cite{lin2022device} 
& Sparse BP 
& Sparse 
& \chruezli
& $\approx$ 
& \chruezli~ \\

TinyTTA~\cite{jia2024tinytta} 
& Early-exit TTA 
& Shallow 
& \chruezli
& N/A
& \chruezli~ \\

AIfES~\cite{wulfert2024aifes}
& Full BP 
& Full 
& \haeggli
& $\approx$* 
& \welleli$^{\ddagger}$  \\

\midrule
\rowcolor{gray!12}
BioTrain 
& Full BP 
& Full 
& \haeggli
& $\approx$ 
& \haeggli~ \\

\bottomrule
\end{tabular}%
}

\vspace{2pt}
\raggedright\footnotesize
$^{\dagger}$ Accuracy relative to offline full-network backpropagation training.\\
$^{\ddagger}$ Supports on-device training but is currently limited to very small CNN and DNN models due to the lack of tiling and memory orchestration.\\
* Only tested on simple MNIST and IRIS datasets.
\end{table}

\textcolor{black}{
Early research on on-device learning primarily addressed memory constraints by avoiding full network adaptation. To reduce activation and gradient storage, these works restricted training to shallow or sparse configurations \cite{ren2021tinyol, profentzas2022minilearn, lin2022device, jia2024tinytta}. As shown in \Cref{tab:related_ondevice_training}, TinyOL \cite{ren2021tinyol} updates only the final classification layer, while MiniLearn \cite{profentzas2022minilearn} and the Tiny Training Engine (TTE) \cite{lin2022device} rely on pruning and sparse \gls{bp} to reduce memory footprints. Similarly, TinyTTA \cite{jia2024tinytta} employs memory-aware early-exit ensembles for lightweight adaptation. 
To further limit activation storage, these approaches typically operate with a batch size of one. 
However, this comes at the cost of a reduced achieved final accuracy.
}

\textcolor{black}{
Crucially, these design choices are not driven by computational infeasibility, as full backpropagation is viable for a wide range of small-scale networks on edge devices \cite{mei_ultra-low_2025}. Instead, the fundamental limitation lies in memory management during training. Supporting full backpropagation on embedded platforms requires storing intermediate activations and gradients that frequently exceed scratchpad memory capacity, making compiler-driven tiling and static memory allocation essential. The only approach that partially supports full BP with an automatic compiler infrastructure to produce end-to-end code is AIfES~\cite{wulfert2024aifes}. However, it lacks systematic mechanisms for tiling and memory orchestration, effectively restricting deployments to small networks that fit entirely within on-chip memory. BioTrain addresses all these limitations by building on Deeploy’s \cite{scherer_deeploy_2024} compiler-supported tiling and static memory allocation, and by combining gradient accumulation with \gls{gn}~\cite{wu_group_2018} to enable batch sizes larger than one, thereby improving the robustness of edge learning algorithms under tight memory budgets and maintaining the same accuracy of offline training.
}

%% file: text/03-methods.tex
\section{BioTrain Framework}
\label{sec:methods}

\begin{figure*}[t]
    \centering
    \includegraphics[width=\textwidth]{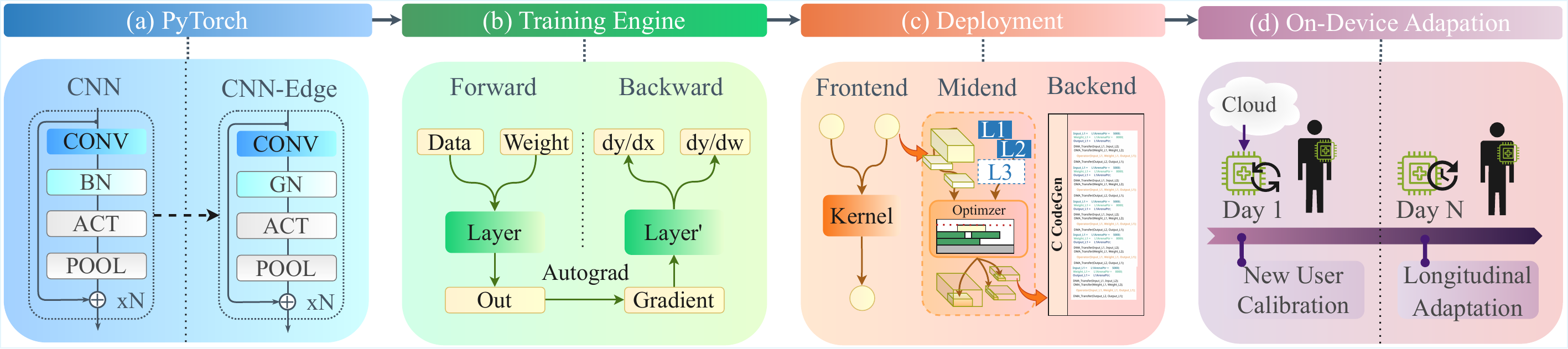}
    \caption{Overview of the BioTrain framework for on-device model adaptation.
(a) The framework interfaces with PyTorch to support common neural network architectures.
(b) The training engine executes forward and backward passes, exposing intermediate activations and gradients through an autograd mechanism to enable parameter updates.
(c) The deployment stack lowers training operators through a frontend--midend--backend compiler pipeline, generating optimized kernels and bare-metal C code with explicit memory management.
(d) In the target on-device learning scenario, a cloud-pretrained model is first initialized and calibrated for a new user, followed by continual on-device adaptation over time.}
    \label{fig:method}
    \vspace{-0.5em}
\end{figure*}


\subsection{Compilation Workflow}
\label{sec:methods:compilation-workflow}
\textcolor{black}{
Figure~\ref{fig:method} summarizes our workflow: in stage~(a), we train the model in PyTorch and replace \gls{bn} with \gls{gn} to avoid cross-sample dependencies and maintain stable optimization under small effective batch sizes. Combined with gradient accumulation, this reduces the peak gradient/activation memory footprint and enables full end-to-end training within \gls{mcu} memory budgets.
}
%
\textcolor{black}{
In stage~(b), we export the trained model via an enhanced ONNX Runtime training API that synthesizes missing gradient subgraphs for unsupported operators, yielding a complete training graph with end-to-end \gls{bp}.
}
%
%
%
\textcolor{black}{
In stage~(c), we compile this ONNX training graph to a target executable by extending Deeploy to support \gls{bp}, generating code for forward and backward passes, and orchestrating memory allocation.
}
\textcolor{black}{
Finally, stage~(d) covers in-field operation: the device performs a Day-1 subject calibration upon first use and then runs periodic on-device fine-tuning to preserve accuracy under longitudinal and temporal distribution shifts in the biosignals.
}

\subsection{On-Chip Memory Reduction for \gls{CNN} Training}
\label{sec:methods:on-chip-memory-reduction}

The first challenge addressed is reducing the peak memory footprint of end-to-end training. In biosignal workloads, typical mini-batches range from 8 to 32. These lead to a proportional increase in activation memory compared to inference, making on-chip training prohibitive under tight memory budgets. While reducing the batch size minimizes the memory usage, it often slows convergence and destabilizes optimization, leading to lower accuracy.
To address this, we employ gradient accumulation, which executes multiple single-sample forward/backward passes and applies weight updates only after reaching the target effective batch size, thereby retaining the optimization benefits of larger batches with a low peak memory footprint.
However, from a compiler perspective, \gls{bn} requires cross-sample reductions to compute batch-level statistics, introducing global synchronization that is incompatible with gradient accumulation. 
\textcolor{black}{We therefore replace \gls{bn} with \gls{gn}, since it has been shown to provide stable optimization when batch sizes are small~\cite{wu_group_2018}, and it depends only on per-sample statistics, avoiding cross-sample synchronization.}
This enables full-network fine-tuning with gradient accumulation under strict on-chip memory constraints.

\begin{figure}[b]
    \centering
    \includegraphics[width=1\linewidth]{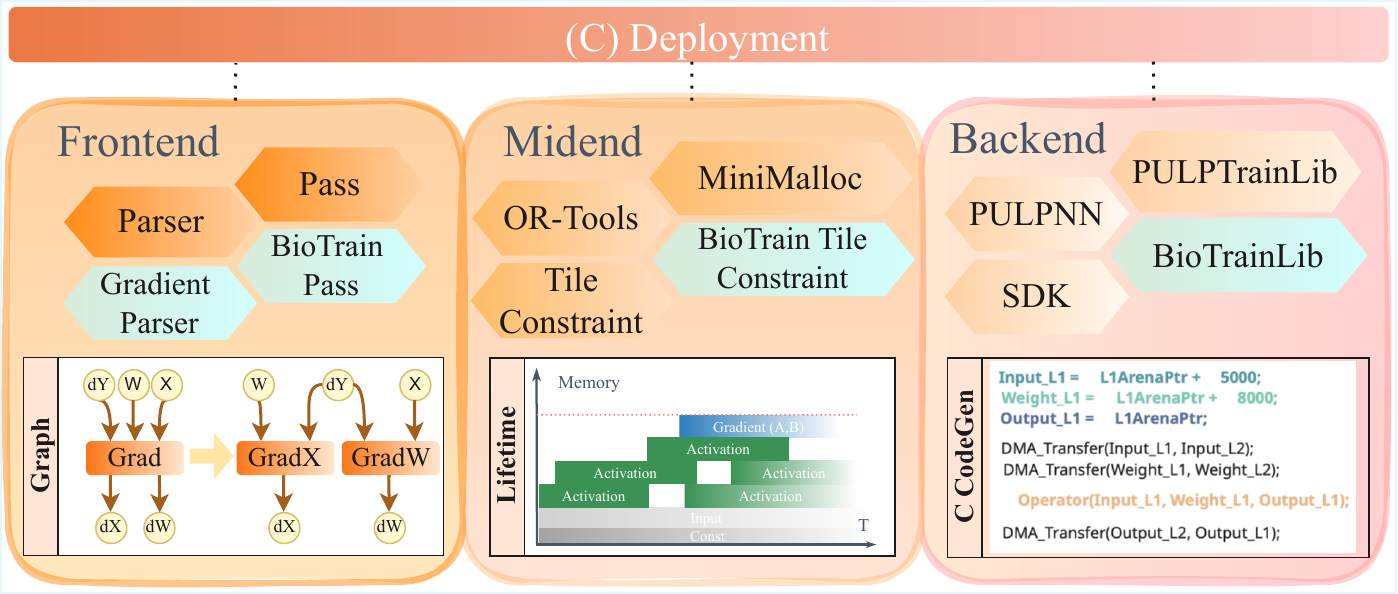}
    \caption{Overview of the deployment stack for end-to-end on-\gls{mcu} training. In light blue, our novel inserted blocks.}
    \label{fig:method2}
    \vspace{-1.8em}
\end{figure}

\subsection{Deployment Stack for End-to-End On-\gls{mcu} Training}

\label{sec:methods:tiling-constraints}
The second challenge addressed is the \textcolor{black}{development} of a deployment stack that produces a final, runnable executable for the target \gls{mcu}, encompassing the complete end-to-end training pipeline in FP32. To this end, we extend Deeploy with three main contributions, summarized in Fig.~\ref{fig:method2}.

\subsubsection{Frontend}
We enhance the frontend to support \gls{CNN} \gls{bp} operators, including \gls{gn}.
We further introduce two frontend passes:
(1) gradient decomposition, which separates parameterized operator gradients into input and weight gradients (i.e., gradient-x and gradient-w) to accommodate different tiling constraints; and
(2) normalization decomposition, which splits normalization operators into statistic computation (i.e., mean and standard deviation) and normalization computation, for handling long biosignals that cannot be fully reduced in L1 memory.

\subsubsection{Midend}
We extend Deeploy's midend to support training graphs, including \gls{bp} and weight-update subgraphs. Deeploy's midend is responsible for tiling and memory allocation across all nodes of the input graph. Given kernel-specific tiling constraints, it uses OR-Tools to solve a constrained optimization problem that determines a scratchpad-aware tiling configuration with high on-chip reuse and minimal off-chip transfers.
The framework generalizes this mechanism to end-to-end training by introducing tiling constraints tailored to \gls{CNN} gradient kernels. Since biosignal workloads typically feature long temporal dimensions and relatively few channels, we adopt temporal tiling as the primary method for orchestrating data movement and reuse within the scratchpad.

For convolution weight gradients, we denote by $T_k$ the temporal tile processed at step $k$. Within each tile, an \texttt{im2col} transformation is applied locally and computes a partial contribution to the weight gradient,
$
\nabla W^{(k)} = \mathrm{im2col}(X_{T_k})^\top \cdot dY_{T_k},
$
which is accumulated into a persistent weight-gradient buffer to form the final $\nabla W$ across all tiles.
For convolution input gradients, $dX$ is tiled temporally, and the corresponding $dY$ tiles are extended with a kernel-dependent halo, ensuring non-overlapping $dX$ tiles and efficient DMA-compute pipelining.
We manage normalization gradients by employing a cross-tile Welford aggregation scheme that computes global mean and variance from tile-local statistics, avoiding the need to retain full-length activations in scratchpad memory. Other operators, including pooling and activation-gradient kernels, follow Deeploy's default tiling strategies.

\subsubsection{Backend}
We extend the Deeploy backend to support \gls{CNN} gradient kernels, with a new library that includes basic convolution gradient operators adapted from PULPTrainLib~\cite{orailoglu_pulp-trainlib_2022}, as well as implemented normalization, pooling, and activation gradient kernels. As PULPTrainLib does not natively support tiling, we introduce modifications to ensure compatibility between its gradient kernels and Deeploy's tiling mechanism. For the forward path, we primarily reuse existing optimized kernels from the Deeploy PULP platform backend and the PULP-NN library.

\subsection{Evaluation Scenarios}
\begin{figure}[tbp]
    \centering
    \includegraphics[width=1.03\linewidth]{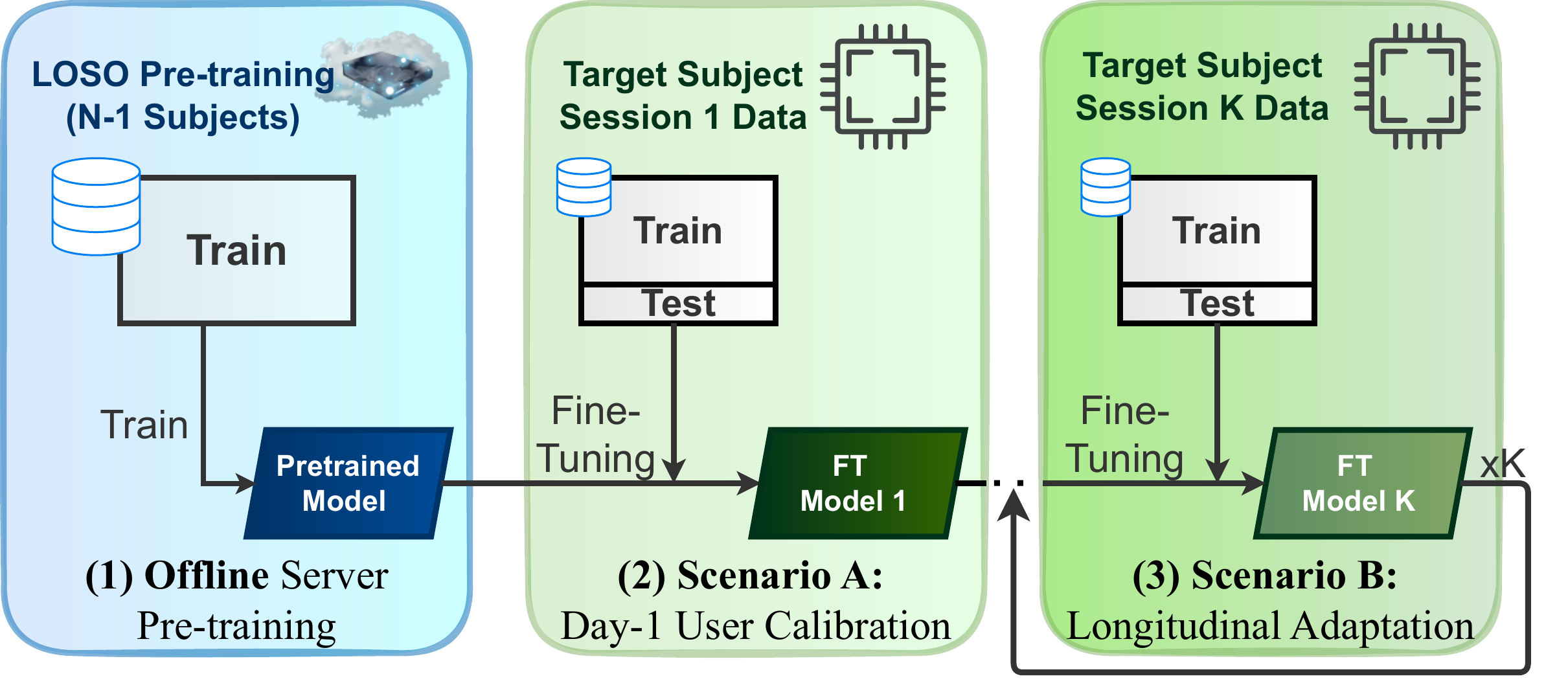}
    \vspace{-1em}
\caption{Overview of the evaluation protocols. The pipeline includes: 
(1) Offline Server Pre-training using a \gls{loso} scheme ($N-1$ subjects); 
(2) \textbf{Scenario A}: Day-1 Calibration. 
(3) \textbf{Scenario B}: Longitudinal Adaptation.}
    \label{fig:evaluation_protocols}
\end{figure}

To systematically evaluate the robustness of fine-tuning methods on edge devices, we provide infrastructure for two scenarios, as illustrated in Fig.~\ref{fig:evaluation_protocols}. 
\glsreset{loso}

\textbf{Scenario A: Day-1 Calibration.}
This protocol simulates initial system deployment to address the cold-start problem for unseen users. \textcolor{black}{We assume the existance of a backbone model obtained offline} via Global \gls{loso}
 Pre-training on $N-1$ subjects. For the target unseen subject, we simulate an initial deployment by fine-tuning the global model on the training set of the subject's first session (80\% of session $S_1$) and evaluating on the test set (20\% of session $ S_1$). \textcolor{black}{This corresponds to an initial calibration session on the subject where ground-truth labels are obtained using a reference setup or supervised protocol (e.g., clinical devices or controlled tasks) while the wearable device collects data.}

\textbf{Scenario B: Longitudinal Adaptation.} We then evaluate a longitudinal adaptation protocol through sequential incremental learning. In this scenario, the model calibrated on the target subject's session $S_1$ (as described in Scenario A) serves as the starting point for continuous evolution. As we collect new data from the user across subsequent sessions ($S_2, S_3, \dots, S_n$), the model continuously refines its tracking of physiological shifts. At each session, we fine-tune on the first 80\% of the data (in temporal order) and test on the remaining 20\%. This simulates continuous refinement to track physiological drift. \textcolor{black}{In a real-world scenario, this corresponds to periodic re-calibration sessions where ground-truth measurements are again collected using reference equipment and supervised procedures to update the on-device model.}

%% file: text/04-results.tex
\section{Experimental Setup}

\begin{table}[!tpb]
\centering
\caption{Summary of the \gls{eeg} and \gls{eog} datasets and models.}
\label{tab:dataset_summary}
\renewcommand{\arraystretch}{1.0}
\setlength{\tabcolsep}{6pt}
\begin{tabular}{lcc}
\toprule
\textbf{Property} & \textbf{\gls{eeg} Dataset} & \textbf{\gls{eog} Dataset} \\
\midrule
Subjects & $5$ & $5$ \\
Sessions per subject & $4$ & $2$ \\
Signal channels & $8$ & $3$ ($L$/$R$/$C$) \\
Classification task & $2$ ($Tongue$ vs. $Rest$) & $11$ \\
Trial duration & $3.8$ $s$ ($T=1900$) & $2.0$ $s$ ($T=1000$) \\
Backbone model & MI-BMINet & EpiDeNet \\
Model parameters & $7.9\,k$ & $4.1\,k$ \\
\bottomrule
\end{tabular}
    \label{table:datasets}

\end{table}

\subsection{Hardware, Biosignal Datasets, and Models}

The hardware evaluation is conducted on the GAP9 \gls{mcu} platform, which integrates a programmable 9-core RISC-V compute cluster and an additional core functioning as the system controller. 
GAP9 features a shared 128\,kB user-managed L1 SRAM and 1.5\,MB L2 SRAM.
Data transfers between L2 and L1 are performed via a programmable DMA engine, enabling overlap between computation and memory transfers. Power consumption is measured using the  Nordic Semiconductor's Power Profiler Kit II.

We evaluate BioTrain across two biosignal modalities, \gls{eeg} and \gls{eog}, using data collected and models deployed via the GAPses~\cite{frey_gapses_2025} platform, which is already built on the GAP9 architecture. The dataset characteristics and model configurations are summarized in Table~\ref{table:datasets}.
For the \gls{eeg} modality, we use Dataset~B from~\cite{mei_ultra-low_2025}, which consists of recordings from $5$ subjects, each collected over $4$ sessions. The task is a binary classification between tongue motor movement and rest. Signals are acquired from $8$ \gls{eeg} channels at a sampling rate of $500\,\mathrm{Hz}$, with each trial lasting $3.8\,\mathrm{s}$ ($T=1900$ samples). The preprocessing pipeline applies a $50\,\mathrm{Hz}$ notch filter followed by a $0.5$--$100\,\mathrm{Hz}$ bandpass filter. We adopt MI-BMINet as the classification backbone, following the original \gls{eeg} study that introduced the dataset. MI-BMINet is a lightweight \gls{CNN} architecture tailored for \gls{eeg} classification and employs spatial--temporal convolutions and depthwise separable convolutions to minimize computational and memory overhead. The model contains $7.9\,\mathrm{k}$ parameters, making it well-suited for on-chip training under tight memory constraints.

For the \gls{eog} modality, we use data collected from the GAPses smart glasses platform~\cite{frey_gapses_2025}, comprising recordings from $5$ subjects across $2$ sessions. Signals are acquired using $3$ dry electrodes positioned on the left and right nose pads and the nasal bridge (center), while bias and reference electrodes are placed on the right and left mastoids, respectively. From these channel measurements, two logical \gls{eog} channels, horizontal and vertical, are derived as
$V_H = V_R - V_L$ and $V_V = V_C - \frac{V_R + V_L}{2}$
following the standard formulation in the original study. Signals are sampled at $500\,\mathrm{Hz}$, with each trial lasting $2.0\,\mathrm{s}$ ($T=1000$ samples). The preprocessing pipeline includes a $50\,\mathrm{Hz}$ notch filter, a $0.5$--$40\,\mathrm{Hz}$ bandpass filter, and a moving average filter with a $2\,\mathrm{s}$ window applied for real-time de-trending, consistent with the original GAPses processing pipeline. The task is an $11$-class eye-movement classification problem. Following~\cite{frey_gapses_2025}, we use EpiDeNet as the classification backbone, a compact \gls{CNN} with $4.1\,\mathrm{k}$ parameters.

\subsection{Evaluation Protocols}

\begin{figure}[t]
\centering

\begin{minipage}{\linewidth}
    \centering
    \footnotesize
    \captionof{table}{Average classification accuracy (\%) under Day-1 Calibration and Longitudinal Adaptation.}
    \label{tab:acc_summary}
    \setlength{\tabcolsep}{6pt}
    \renewcommand{\arraystretch}{1.1}

\setlength{\tabcolsep}{3pt}\begin{tabular*}{\linewidth}{l@{\extracolsep{\fill}}cccc}
\toprule
\textbf{Task}
& \multicolumn{4}{c}{\textbf{Avg. Acc. (\%, Mean $\pm$ Std.)}} \\
\cmidrule(lr){2-5}
& \textbf{No-FT}
& \textbf{\gls{lp}}
& \textbf{Full-FT}
& \textbf{Edge-FT} \\
\midrule
\gls{eeg} (Subj.)
& $50.8 \pm 8.8$
& $74.8 \pm 11.4$
& $80.0 \pm 11.4$
& \cellcolor[HTML]{F2F2F2}\textbf{86.4} $\pm 7.9$ \\
\addlinespace[0.25em]
\gls{eeg} (Sess.)
& $49.8 \pm 3.8$
& $76.2 \pm 7.6$
& $78.7 \pm 7.3$
& \cellcolor[HTML]{F2F2F2}\textbf{83.9} $\pm 13.3$ \\
\addlinespace[0.25em]
\gls{eog} (Subj.)
& $78.1 \pm 18.1$
& $83.3 \pm 13.0$
&\cellcolor[HTML]{F2F2F2}\textbf{88.7} $\pm 10.3$
& $87.7 \pm 10.4$ \\
\addlinespace[0.25em]
\gls{eog} (Sess.)
& $84.1 \pm 11.2$
& $86.5 \pm 10.7$
& \cellcolor[HTML]{F2F2F2}\textbf{89.1} $\pm 8.8$
& $87.2 \pm 7.6$ \\
\bottomrule
\label{offline-table}
\end{tabular*}
\end{minipage}

\vspace{-0.6em}
\begin{minipage}{\linewidth}
    \centering
    \includegraphics[width=\linewidth]{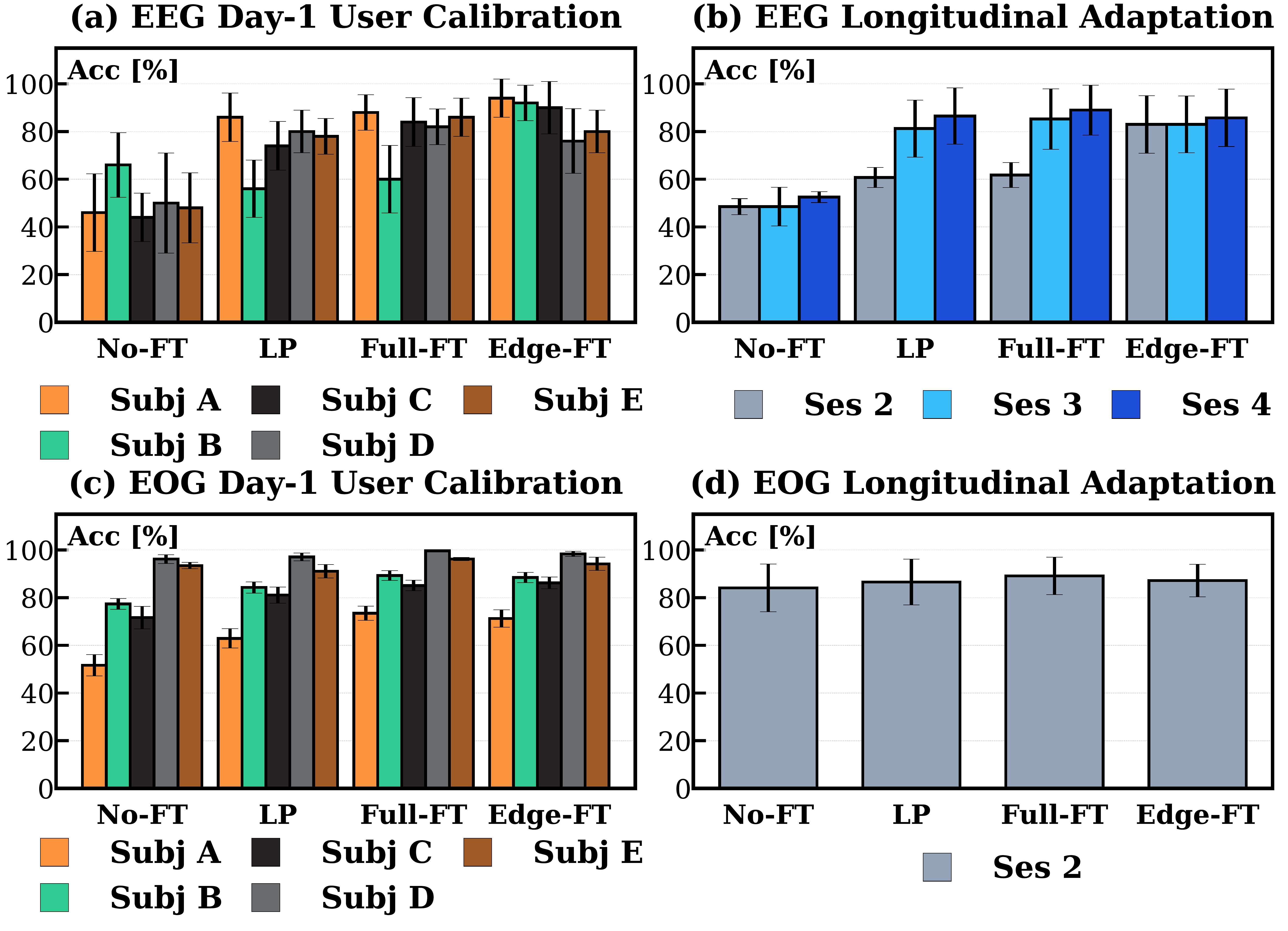}
    \captionof{figure}{Comparison of fine-tuning strategies across Day-1 Calibration (a, c) and Longitudinal Adaptation (b, d) for \gls{eeg} (top) and \gls{eog} (bottom). \gls{edgeft} replaces \gls{bn} with \gls{gn} to enable memory-efficient on-device training.}
    \label{fig:offline}
\end{minipage}
\vspace{-0.8em}
\end{figure}

For both modalities (\gls{eeg} and \gls{eog}) under both scenarios (Day-1 Calibration and Longitudinal Adaptation), we evaluate four adaptation strategies: \gls{noft}, \gls{lp} (only training the last classification layer), \gls{fullft} (using the original architectures with \gls{bn} layers), and the proposed edge-optimized \gls{edgeft}, which replaces \gls{bn} with \gls{gn} to improve training stability under small-batch, on-device settings. For cloud pre-training, we use the \gls{adamw} optimizer with a batch size of 64, a learning rate of $1\times10^{-3}$, and 40 training epochs. To accommodate the transition from cloud-scale resources to edge-constrained environments, on-device adaptation is performed using a more memory-efficient optimization setup. Specifically, we switch to \gls{sgd} with momentum (0.9) and cosine annealing, using a reduced batch size of 8 and a learning rate of $5\times10^{-3}$, with weight decay set to $1\times10^{-3}$. The number of fine-tuning epochs is set to 30. All experiments are repeated five times with different random seeds, and results are reported as mean and standard deviation.


\section{Experimental Results}

\subsection{Adaptation Accuracy}

Figure~\ref{fig:offline} and Table~\ref{offline-table} summarize the accuracy of the described protocols under the two introduced scenarios for both benchmarks.

\textbf{Scenario~A: Day-1 Calibration.}
Without adaptation, \gls{noft} yields low accuracy for \gls{eeg} (50.8\%) and high variance for \gls{eog} (78.1\% $\pm$ 18.1\%), confirming that pre-trained models cannot generalize to new users. \gls{lp} improves accuracy to 74.8\% (\gls{eeg}) and 83.3\% (\gls{eog}), yet exhibits large cross-subject variability, with \gls{eeg} Subject~B saturating at 56.0\%, indicating classifier-only updates are insufficient to cope with distribution shifts. \gls{fullft} further improves to 80.0\% (\gls{eeg}) and 88.7\% (\gls{eog}), demonstrating the benefit of updating the full network. \gls{edgeft} achieves comparable or better performance, 86.4\% for \gls{eeg} (+6.4\% over \gls{fullft}) and 87.7\% for \gls{eog} while reducing variance, showing that replacing \gls{bn} with \gls{gn} does not compromise adaptation quality.

\textbf{Scenario~B: Longitudinal Adaptation.}
Without adaptation, \gls{noft} degrades to 49.8\% for \gls{eeg} and remains at 84.1\% for \gls{eog}, confirming persistent signal drift across sessions.
\gls{lp} shows adaptation lag for \gls{eeg}, with accuracy dropping to 62.4\% at session~$S_2$ before gradually recovering; such delayed convergence is impractical for wearables. For \gls{eog}, \gls{lp} achieves 86.5\%, offering only marginal improvement.
\gls{fullft} exhibits similar lag for \gls{eeg} (61.6\% at $S_2$) due to fragile representations from weak Day-1 initialization, but achieves the best \gls{eog} accuracy (89.1\%), demonstrating its effectiveness when initial calibration is strong.
\gls{edgeft} maintains stable \gls{eeg} performance from $S_2$ onward (84.8\% at $S_2$, 85.8\% at $S_4$), avoiding the adaptation lag observed with other methods. For \gls{eog}, \gls{edgeft} achieves 87.2\%, comparable to \gls{fullft}.
These results confirm that full-network fine-tuning tracks effectively signal drift, with \gls{edgeft} providing the most consistent longitudinal adaptation across both modalities.
In both scenarios and across both benchmarks, we have demonstrated that our \gls{edgeft} achieves comparable or better accuracy to \gls{fullft}, while outperforming the lighter \gls{lp}.
\vspace{-0.5em}

\begin{table*}[!ht]
\centering
\caption{\textcolor{black}{System-level evaluation of on-device training on GAP9 (FP32, 1.5\,MB L2), comparing LP, Full-FT, and Edge-FT.}}

\label{tab:system_metrics}
\scriptsize
\setlength{\tabcolsep}{3pt}
\renewcommand{\arraystretch}{1.0}
\resizebox{\textwidth}{!}{
\begin{tabular}{l l c c c c c c c c}
\toprule
\textbf{Dataset} & \textbf{Strategy} & \textbf{Acc (Subj/Sess)} & \textbf{Params (Train)} & \textbf{FLOPs} & \textbf{L2 Peak Memory (MB)} & \textbf{Latency (ms)} & \textbf{Power (mW)} & \textbf{Energy (mJ)} & \textbf{Throughput (GFLOPs/s)} \\
\midrule
\multirow{3}{*}{\textbf{\gls{eeg}~\cite{mei_ultra-low_2025}}}
  & \gls{lp}
  & 74.8 / 76.2
  & 0.07\,k
  & 139.2\,M 
  & 0.19 
  & 360.0
  & 45.2 
  & 16.08 
  & 0.38 \\
  & Full-FT
  & 80.0 / 78.7
  & 7.9\,k
  & 416.8\,M
  & 5.36$^\dagger$
  & ---
  & ---
  & ---
  & --- \\
\cmidrule(lr){2-10}
  & \cellcolor[HTML]{F2F2F2}\gls{edgeft}
  & \cellcolor[HTML]{F2F2F2}\textbf{86.4 / 83.9}
  & \cellcolor[HTML]{F2F2F2}7.9\,k 
  & \cellcolor[HTML]{F2F2F2}416.8\,M 
  & \cellcolor[HTML]{F2F2F2}0.67
  & \cellcolor[HTML]{F2F2F2}469.6
  & \cellcolor[HTML]{F2F2F2}43.8
  & \cellcolor[HTML]{F2F2F2}20.16
  & \cellcolor[HTML]{F2F2F2}0.89 \\
\midrule
\multirow{3}{*}{\textbf{\gls{eog}~\cite{frey_gapses_2025}}}
  & \gls{lp}
  & 83.3 / 86.5
  & 0.2\,k 
  & 7.2\,M 
  & 0.11 
  & 34.4 
  & 50.4 
  & 1.76 
  & 0.21 \\
  & Full-FT
  & \textbf{88.7 / 89.1}
  & 4.1\,k
  & 20.8\,M
  & 2.24$^\dagger$
  & ---
  & ---
  & ---
  & --- \\
\cmidrule(lr){2-10}
  & \cellcolor[HTML]{F2F2F2}\gls{edgeft}
  & \cellcolor[HTML]{F2F2F2}87.7 / 87.2
  & \cellcolor[HTML]{F2F2F2}4.1\,k  
  & \cellcolor[HTML]{F2F2F2}20.8\,M 
  & \cellcolor[HTML]{F2F2F2}0.28 
  & \cellcolor[HTML]{F2F2F2}93.6 
  & \cellcolor[HTML]{F2F2F2}48.2 
  & \cellcolor[HTML]{F2F2F2}4.48
  & \cellcolor[HTML]{F2F2F2}0.22 \\
\bottomrule
\end{tabular}
}

\vspace{0.3em}
\footnotesize{$^\dagger$ Exceeds GAP9 L2 capacity (1.5\,MB), thus cannot execute fully on-chip.}

\vspace{-1em}
\label{deeploy-benchmark}
\end{table*}

\subsection{\textcolor{black}{Memory, Performance, Energy Analysis}}

Table~\ref{deeploy-benchmark} reports per-batch (8 samples) system-level performance on GAP9. For \gls{eeg}, \gls{edgeft} scales trainable parameters from 0.07\,k (\gls{lp}) to 7.9\,k and per-batch compute from 139.2\,M to 416.8\,M FLOPs, yet caps peak L2 activation memory at 0.67\,MB---enabling full-network \gls{bp} entirely on-chip. A conventional full-FT with standard batch normalization would require $\sim$5.4\,MB, exceeding GAP9's 1.5\,MB L2. BioTrain sustains 0.89\,GFLOPs/s under 50\,mW; \gls{eog} shows a similar trend (0.28\,MB, 0.22\,GFLOPs/s).

With 40 epochs over 200 samples (8000 samples/session) on a 320\,mAh, 3.7\,V battery, BioTrain affords $\sim$211 \gls{eeg} and $\sim$951 \gls{eog} on-device training sessions, demonstrating practical full-network fine-tuning within wearable-class \gls{mcu} energy and memory budgets.

%% file: text/05-conclusions.tex
\section{Conclusion}

We presented BioTrain, a compiler-assisted on-device adaptation framework enabling full-network fine-tuning within a sub-50\,mW envelope on \gls{mcu} platforms. On \gls{eeg} and \gls{eog} datasets it delivers up to 35\% accuracy gains over non-adapted baselines with 8$\times$ peak memory reduction (5.4\,MB$\to$0.67\,MB), enabling full on-chip \gls{bp} across Day-1 Calibration and Longitudinal Adaptation scenarios. Future work will extend BioTrain to quantized training and additional biosignal modalities.